\documentclass{article}
\usepackage{arxiv}

\usepackage[table,xcdraw]{xcolor}
\usepackage[utf8]{inputenc}
\usepackage[english]{babel}
\usepackage[square,numbers]{natbib}
\usepackage{amsmath}
\usepackage{float}
\usepackage{amsthm}
\usepackage{mathrsfs}
\usepackage{amssymb}
\usepackage{blkarray, bigstrut}
\usepackage{pgfplots}
\usepackage{booktabs}
\usepackage{hyperref}
\usepackage{hhline}

\usepackage{doi}
\usetikzlibrary{spy}
\usepackage{tikz,pgfplots,pgf}
\usetikzlibrary{arrows,shapes,calc}
\usetikzlibrary{positioning}
\usepackage{neuralnetwork}
\usepackage[normalem]{ulem}

\newtheorem{theorem}{Theorem}
\newtheorem{proposition}{Proposition}
\newtheorem{definition}{Definition}
\newtheorem{property}{Property}
\newtheorem{corollary}{Corollary}
\newtheorem{lemma}{Lemma}

 \pgfplotsset{compat=1.18} 

\title{On learning capacities of Sugeno integrals with systems of fuzzy relational equations}

\hypersetup{
pdftitle={On learning capacities of Sugeno integrals with systems of fuzzy relational equations},
pdfsubject={cs.AI},
pdfauthor={Ismaïl Baaj},
pdfkeywords={sugeno integrals, system of fuzzy relational equations, learning},
}

\author{Ismaïl Baaj \\ Univ. Artois, CNRS, CRIL, F-62300 Lens, France \\  \href{baaj@cril.fr}{baaj@cril.fr}}
\begin{document}
\maketitle

\begin{abstract}
In this article, we introduce a method for learning a capacity underlying a Sugeno integral according to training data based on systems of fuzzy relational equations. 
To the training data, we associate two systems of equations: a $\max-\min$ system and a $\min-\max$ system.
By solving these two systems (in the case that they are consistent) using Sanchez's results, we show that we can directly obtain the extremal capacities representing the training data.  By reducing the $\max-\min$ (resp. $\min-\max$) system of equations to subsets of criteria of cardinality less than or equal to $q$ (resp. of cardinality greater than or equal to $n-q$),  where $n$ is the number of criteria, we give a sufficient condition for deducing, from its potential greatest solution (resp. its potential lowest solution), a $q$-maxitive (resp. $q$-minitive) capacity. 
Finally, if these two reduced systems of equations are inconsistent, we show how to obtain the greatest approximate $q$-maxitive capacity and the lowest approximate $q$-minitive capacity, using recent results to handle the inconsistency of systems of fuzzy relational equations.
\end{abstract}


\section{Introduction}

In decision theory, Sugeno integral \cite{Sug74} is commonly used as a qualitative aggregation function \cite{grabisch2010decade} with values on a finite scale $L$. The definition of this integral is based on an set function called capacity, which qualitatively models the importance (or interaction) of subsets of criteria, and is used in many fields \cite{grabisch2016set} such as uncertainty modeling, multicriteria aggregation or in game theory.

In recent years, many approaches, e.g. \cite{abbaszadeh2020machine, beliakov2020robust, AlguesSugeno2015}, have been proposed to learn (in closely related senses) the capacity of a Sugeno integral according to $N$ training data, each training data item being a pair (object, targeted global evaluation).

In this article, we propose an approach for learning a capacity underlying a Sugeno integral according to training data, which is based on systems of fuzzy relational equations. Our goal is as follows: the Sugeno integral defined by a learned capacity should allow us to obtain, for each object in the training data, its targeted global evaluation, or {\it an approximated value in a specific sense}. 
Our approach is based on the fact that the definition of the Sugeno integral of an object defined by a capacity is given by a $\max-\min$ product between a row matrix containing the minimum of the partial evaluations of the object studied with respect to the subsets of criteria, and a column vector whose components are the measures of the subsets of criteria defined by the capacity. The Sugeno integral can also be obtained by a  $\min-\max$ product similarly defined. 

These two expressions of the Sugeno integral allow us to introduce two systems of fuzzy relational equations, one of type $\max-\min$ and a second of type $\min-\max$ for our learning problem. The matrix of each system is constructed from the objects in the training data, while the second member of each system consists of the targeted global evaluations. Under some natural conditions, the solving of each of the two systems, based on the work of Sanchez \cite{sanchez1976resolution}, produces a vector whose components define the measures of the subsets of criteria of a capacity. 

 \noindent
The $q$-maxitive/minitive capacities were introduced in \cite{calvo1998aggregation,mesiar1999k,wu2019k} to reduce the complexity of the computation of the global evaluation of an object by the Sugeno integral. In this article, we show that learning $q$-maxitive (resp. $q$-minitive) capacities can be studied by reducing the $\max-\min$ (resp. $\min-\max$) system to subsets of criteria of cardinality less than or equal to $q$ (resp. of cardinality greater than or equal to $n - q$),  where $n$ is the number of criteria.
To establish a necessary and sufficient condition (Theorem \ref{th:qmaxred}, resp. Theorem \ref{th:qmin}) in order that a $q$-maxitive (resp. $q$-minitive) capacity represents the training data, we begin by giving a construction of such a capacity (Proposition \ref{propg:qmax}) (resp. (Proposition \ref{propg:qprmin})).  In the case where the reduced   $\max-\min$ system  (resp. reduced $\min-\max$ system) is inconsistent, under a natural assumption on the data, we rely on the results of \cite{baaj2024handling} to compute explicitly in Theorem \ref{th:DeltaProp} (resp. Theorem \ref{th:DeltaProp1}), the greatest (resp. lowest) $q$-maxitive (resp. $q$-minitive) capacity  which approximately represents the training data in a specific sense, see  
Corollary \ref{cor:greatestqmax} (resp. Corollary \ref{cor:lowestqmin}).

The article is structured as follows. Section \ref{sec:bg} gives the necessary background on the Sugeno integral and systems of fuzzy relational equations. In Section \ref{sec:learning}, we introduce two systems of fuzzy relational equations that allow us to learn capacities of Sugeno integrals according to training data, and we give a construction of the $q$-maxitive (resp. $q$-minitive) capacities. We illustrate our results on an example. 
In Section \ref{sec:qmaxqmin}, we state our main results on learning $q$-maxitive/minitive capacities.

\section{Background}
\label{sec:bg}
In what follows, we recall the definition of the Sugeno  integral and the main results for solving systems of fuzzy relational equations. 
\subsection{Sugeno integral}
The Sugeno integral framework (see \cite{dubois2001use} for more details) includes:
\begin{itemize}
    \item[-] $\mathcal{C} = \{1,2,\dots,n\}$ : a set of $n$ criteria,  
    \item[-] $L = \{ \xi_1 = 0 , \xi_2 , \dots  , \xi_l = 1 \}$ :  an evaluation scale (a totally ordered finite set $\xi_1 = 0 < \xi_2 < \dots  < \xi_l = 1$) or $L = [0  , 1]$. 
    \item[-]~$\mu: 2^\mathcal{C} \rightarrow L$  a capacity, i.e., a set function such that $\mu(\emptyset)=0$,$\mu(C)=1$ and $A \subseteq B \Longrightarrow \mu(A) \leq \mu(B)$. The conjugate capacity $\mu^c$ is defined by  $\mu^c(A) = \iota(\mu(\overline{A}))$ with    $\iota : L \rightarrow L : \xi_i \mapsto \xi_{l -i + 1}$, or  
$\iota : [0 , 1] \rightarrow [0 , 1] : t \mapsto 1 - t$.
    \item[-]  $x = [x_i]\in  L^{n \times 1}$ : an object where each $x_i$ is a partial evaluation of the object $x$ according to the criterion $i$.
\end{itemize}

\noindent The Sugeno integral is given by two equivalent formulas \cite{marichal2000sugeno,dubois2001use} (other equivalent formulas exist): 
\begin{align}
    S_\mu(x)  &= \max_{A \in 2^\mathcal{C}}\, \min(\min_{i \in A} x_i , \mu(A)) \label{eq:sugenoformulas-maxmin}\\
    &= \min_{A \in 2^\mathcal{C}}\, \max(\max_{i \in \overline A} x_i, \mu(A)).  \label{eq:sugenoformulas-minmax}
\end{align}
where $\overline A$ denotes the complementary of the subset $A$ of $\mathcal C$.
\noindent
The following two notions have been introduced in \cite{calvo1998aggregation,mesiar1999k,wu2019k}:
\begin{definition}
   Let $\mu : 2^{\mathcal C} \rightarrow L$ a capacity and $q\in \{1 , 2 , \dots , n\}$ where $n = \mid \mathcal C\mid $.
    \begin{enumerate}
        \item $\mu$ is said to be $q$-maxitive iff for all subsets $X \subseteq \mathcal C$
verifying $\mid X \mid > q$, we have $\mu(X) = \max_{Y \subset X , \mid Y \mid \leq q} \mu(Y)$.
\item $\mu$ is said to be  $q$-minitive iff for all subsets  $X \subseteq \mathcal C$
verifying  $\mid X \mid < (n - q)$, we have $\mu(X) = \min_{Y \supset X , \mid Y \mid \geq (n - q)} \mu(Y)$.

\end{enumerate}
\end{definition}
\noindent

It is easy to check that a capacity $\mu$ is $q$-maxitive iff its conjugate $\mu^c$ is $q$-minitive.  Any capacity $\mu: 2^{\mathcal C}  \rightarrow L$ is (trivially) $n$-maxitive and $n$-minitive. 
Moreover, the two formulas (\ref{eq:sugenoformulas-maxmin}) and (\ref{eq:sugenoformulas-minmax}) become:

\begin{lemma}\label{qintS}
Let $\mu : 2^{\mathcal C} \rightarrow L$ a capacity,  $q\in \{1 , 2 , \dots , n\}$ ,  $n = \mid\mathcal{C}\mid$ and $x =\begin{bmatrix}x_i\end{bmatrix} \in L^{n \times 1}$.
\begin{enumerate}
    \item If  $\mu$ is $q$-maxitive, then:
\begin{equation*}\label{eq:qmaxintS}
S_\mu(x) = \max_{A\in 2^{\mathcal C}\backslash\{\emptyset\}, \mid A \mid \leq q} \min(\min_{i \in A} x_i , \mu(A)).
\end{equation*}
\item If $\mu$ is $q$-minitive, then we have: 
\begin{equation*}\label{eq:qminintS}
S_\mu(x) = \min_{A\in 2^{\mathcal C}\backslash\{\mathcal C\}, \mid A \mid \geq n - q} \max(\max_{i \in \overline A} x_i , \mu(A)).
\end{equation*}
\end{enumerate}
\end{lemma}
The $q$-maxitive and $q$-minitive capacities allow us to compute the global evaluation of an object $x$ by considering fewer subsets of criteria than in the general case.

\subsection{Solving systems of max-min fuzzy relational equations}

\noindent A system of fuzzy relational equations of type $\max-\min$ based on a matrix $A=[a_{ij}] \in L^{n\times m}$ of size $(n,m)$ and a vector $b=[b_{i}] \in L^{n\times 1}$ of $n$ components, is of the following form:
\begin{equation}\label{eq:sysmaxmin}
    (S): A \Box_{\min}^{\max} x = b,
\end{equation}
\noindent where $x = [x_j]_{1 \leq j \leq m} \in L^{m\times 1}$ is an unknown vector with $m$ components and the operator $\Box_{\min}^{\max}$ is the matrix product which uses the t-norm $\min$ as the product and the function $\max$ as the addition. 

\noindent From the vector \begin{equation}\label{eq:gr:sol}
    e = A^t \Box_{{\rightarrow_G}}^{\min} b,
\end{equation}
\noindent where $A^t$ is the transpose of $A$ and the matrix product $\Box_{{\rightarrow_G}}^{\min}$ use the Gödel implication  $\rightarrow_G$ defined by: 
\begin{equation}\label{eq:godelimp}
    x \rightarrow_G y = \begin{cases}
    1 & x \leq y\\
    y & \text{otherwise}.
    \end{cases}
\end{equation}
as the product and the function $\min$ as the addition, Sanchez \cite{sanchez1976resolution} showed the following equivalence:
\begin{equation*}\label{eq:consistemaxmin}
     \text{The system $(S)$ is consistent }\Longleftrightarrow A \Box_{\min}^{\max} e = b. 
\end{equation*}
If the system $(S)$ is consistent, the vector $e$ is its greatest solution \cite{sanchez1976resolution}.

\subsection{Solving systems of min-max fuzzy relational equations}

\noindent A system of fuzzy relational equations of type $\min-\max$ can be defined similarly. Given a matrix $\Gamma=[\gamma_{ij}] \in L^{n\times m}$ of size $(n,m)$ and a vector $\beta=[\beta_{i}] \in L^{n\times 1}$ of $n$ components, we define a system of $\min-\max$ fuzzy relational equations by:
\begin{equation}\label{eq:sysminmax}
    (\Sigma): \Gamma \Box_{\max}^{\min} x = \beta,
\end{equation}
\noindent where $x = [x_j]_{1 \leq j \leq m} \in L^{m\times 1}$ is an unknown vector with $m$ components and the operator $\Box_{\max}^{\min}$ is the matrix product which uses the function $\max$ as the product and the function $\min$ as the addition. 

\noindent From the vector \begin{equation}\label{eq:lt:sol}
    f = \Gamma^t \Box_{\epsilon}^{\max} \beta,
\end{equation}
\noindent where $\Gamma^t$ is the transpose of $\Gamma$ and the matrix product $\Box_{{\epsilon}}^{\max}$ uses the epsilon product (denoted $\epsilon$) defined by: 
\begin{equation}\label{eq:epsilonprod}
    x \epsilon y = \begin{cases}
    y & x < y\\
    0 & x \geq y.
    \end{cases}
\end{equation}
as the product and the  function $\max$ as the addition, we have the following equivalence:
\begin{equation*}\label{eq:consisteminmax}
     \text{The  system $(\Sigma)$ is consistent }\Longleftrightarrow \Gamma \Box_{\max}^{\min} f = \beta. 
\end{equation*}
If the system $(\Sigma)$ is consistent, the vector $f$ is its lowest solution \cite{sanchez1976resolution}.
\section{Learning capacities for Sugeno integrals according to data with systems of fuzzy relational equations}
\label{sec:learning}
In this section, we introduce our method for learning capacities for Sugeno integrals based on training data using systems of fuzzy relational equations. We consider a set of $N$ training data $\{ (x^{(1)}, \alpha^{(1)}), (x^{(2)}, \alpha^{(2)}), \dots, (x^{(N)}, \alpha^{(N)}) \}$. For $k=1,2,\dots,N$, each pair $(x^{(k)}, \alpha^{(k)})$ consists of an object  $x^{(k)} = \begin{bmatrix}x^{(k)}_i\end{bmatrix}  \in L^{n \times 1}$     which contains the partial evaluation $x_i^{(k)}$ of the object $x^{(k)}$ according to the criterion $i$, and the targeted global evaluation of the object, which is denoted by $\alpha^{(k)} \in L$.

\subsection{Constructing two systems of fuzzy relational equations from training data}
From the data $(x^{(k)} , \alpha^{(k)})_{1 \leq k \leq N}$, we construct two matrices and a vector:
\begin{itemize}
    \item[-] $M = [m_{k ,A}]_{ 1 \leq k \leq N , A\in 2^\mathcal{C} \backslash  \{\emptyset\}} \text{ where }  m_{k , A} = \min\limits_{i \in A}  x^{(k)}_i$,
    \item[-] $\Gamma = [\gamma_{k ,A}]_{ 1 \leq k \leq N , A\in 2^\mathcal{C} \backslash  \{\mathcal{C}\}} \text{ where }  \gamma_{k , A} = \max\limits_{i \in \overline A}  x^{(k)}_i$,
    \item[-] $\alpha = [\alpha^{(k)}]\in L^{N \times 1}$: a vector containing the targeted global evaluations of the objects.
\end{itemize}

We form the following system of fuzzy relational equations of type $\max-\min$ from the matrix $M$ and the vector $\alpha$:
\begin{equation}\label{eq:sysLearnS}
    (S) : M \Box_{\min}^{\max} X = \alpha
\end{equation}
\noindent where $X = \begin{bmatrix}
     \xi_A
 \end{bmatrix}_{A \in 2^\mathcal{C}\backslash \{\emptyset\}}$ is an unknown vector and $e = M^t \Box_{\rightarrow_G}^{\min}\alpha$ is the potential greatest solution of the system  $(S)$, see (\ref{eq:gr:sol}). 
 We form the following system of fuzzy relational equations of type $\min-\max$ from the matrix $\Gamma$ and the vector $\alpha$:
\begin{equation}\label{eq:sysLearnSigma}
    (\Sigma) : \Gamma \Box_{\max}^{\min} X = \alpha
\end{equation}
\noindent where $X=\begin{bmatrix}
     \xi_A
 \end{bmatrix}_{A \in 2^{\mathcal{C}}\backslash\{\mathcal{C}\}}$ is an unknown vector and $f = \Gamma^t \Box_{\epsilon}^{\max}\alpha$ is the  potential lowest solution of the system  $(\Sigma)$, see (\ref{eq:lt:sol}). 
\subsection{Preliminary results}
The potential extremal solutions of the two systems $(S)$ and $(\Sigma)$ viewed as set functions $e : 2^{\mathcal{C}} \backslash \{ \emptyset \} \rightarrow L$ and $f : 2^\mathcal{C}\backslash\{\mathcal{C}\} \rightarrow L$ are increasing:
\begin{lemma}\mbox{} \label{lemma:croissance}
\begin{enumerate}
    \item $\emptyset \subset A \subseteq   B   \Longrightarrow \,  e(A) \leq e(B)$. 
    \item $A \subseteq   B \subset \mathcal{C}  \Longrightarrow \,  f(A) \leq f(B)$.
\end{enumerate}
\end{lemma}
\noindent
These two implications can be deduced from the properties of Gödel's implication  $\rightarrow_G$, see (\ref{eq:godelimp})  and those of the epsilon product $\epsilon$, see  (\ref{eq:epsilonprod}):  $\forall x , x'\in L$ such that $x \leq x'$, we have  
$\forall y\in L,\, x' \rightarrow_G y \leq x \rightarrow_G y$ and   $x' \epsilon y \leq x \epsilon y$.

\noindent
For all $\emptyset \subset A \subseteq \mathcal C$ and  
$ B \subset  \mathcal C$, we put:
\begin{itemize}
    \item $J_A=\{k \in \{1 , 2 , \dots , N\}\mid  \min\limits_{i \in A} x_i^{(k)} >  \alpha^{(k)}\}$,
    \item $K_B = \{k \in \{1 , 2 , \dots , N\} \mid  \max\limits_{i \in \overline B} x_i^{(k)} <  \alpha^{(k)}\}$.
\end{itemize}
 With these notations, we have:
\begin{lemma}\mbox{} \label{lemma:calculs}
\begin{enumerate}
    \item $e(A) = \min\limits_{k\in J_A} \alpha^{(k)},  \text{ with the convention: } \min_\emptyset = 1$. 
    \item $f(B) = \max\limits_{k\in K_B} \alpha^{(k)},  \text{ with the convention: } \max_\emptyset = 0$.
\end{enumerate}
\end{lemma}

\noindent
In particular, we have:
\begin{itemize}
    \item $e(C) = 1 \Longleftrightarrow \forall k\in\{1 , 2 , \dots , N\}, \,\, \min\limits_{i \in \mathcal C} x_i^{(k)}  \leq \alpha^{(k)}$,
    \item $f(\emptyset) = 0 \Longleftrightarrow  \forall k\in\{1 , 2 , \dots , N\},  \,\,  \alpha^{(k)} \leq \max\limits_{i \in \mathcal C} x_i^{(k)}$.
\end{itemize}

\subsection{Learning capacities}
\noindent
We construct two set functions from the potential extremal solutions $e$ and $f$ of the systems $(S)$ and $(\Sigma)$ defined from the training data, see (\ref{eq:sysLearnS}) and (\ref{eq:sysLearnSigma}), as follows:
\noindent  

\begin{equation}\label{eq:mue}
    \mu_e(A) = \left\lbrace{\begin{array}{rrl}e(A) & \quad \text{if } \quad A \not=\emptyset\\
0 & \quad \text{if } \quad A =\emptyset \end{array}},
\right.
\end{equation}
\begin{equation}\label{eq:muf}
    \mu_f(A) = \left\lbrace{\begin{array}{rrl}f(A) & \quad \text{if } \quad A \subset \mathcal{C}\\
1 & \quad \text{if } \quad A =\mathcal{C}\end{array}}.\right.
\end{equation}
To guarantee that $\mu_e,\mu_f$ are capacities which are compatible with the training data, we want the systems $(S)$ and $(\Sigma)$ to be consistent (i.e., to have solutions) and we want $\mu_e(\mathcal{C}) = e(\mathcal{C})= 1$ and $\mu_f(\emptyset) = f(\emptyset) = 0$.
We establish the following result:
\begin{theorem}\label{th1}
The following three conditions are equivalent:
\begin{enumerate}
    \item  There exists a capacity  $\mu : 2^\mathcal{C} \rightarrow L$  which represents the training data  i.e., a capacity such that 
$\forall k\in \{1 , 2 , \dots , N\}, \text{ we have } S_\mu(x^{(k)}) = \alpha^{(k)}$, 
        \item The system $(S) : M \Box_{\min}^{\max} X = \alpha$ is consistent and  $e(\mathcal{C})= 1$,
        \item The system  $(\Sigma) : \Gamma \Box_{\max}^{\min} X = \alpha$ is consistent and $f(\emptyset) = 0$.
\end{enumerate}
\end{theorem}
\noindent
{\it Sketch of the proof:} for any capacity $\mu : 2^{\mathcal C} \rightarrow L$, we put: $v : 2^{\mathcal C}\backslash\{\emptyset\} \rightarrow L : A \mapsto v(A) = \mu(A)$ and $w : 2^{\mathcal C}\backslash\{\mathcal C\} \rightarrow L : A \mapsto w(A) = \mu(A)$. 
Therefore, it results from the formulas (\ref{eq:sugenoformulas-maxmin}) and (\ref{eq:sugenoformulas-minmax}) that we have:
\begin{align*}
   \forall k\!\in\!\{1 , 2 , \dots , N\}, S_\mu(x^{(k)})\!=\alpha^{(k)}&  \! \Longleftrightarrow\!\!M\Box_{\min}^{\max} v = \alpha  \\
    &\!\Longleftrightarrow \Gamma\Box_{\max}^{\min} w = \alpha.
\end{align*}

\begin{corollary}\label{cor}
If the equivalent conditions of Theorem \ref{th1} are verified, we have:
\begin{itemize}
    \item  $\mu_e$  is a capacity that represents the training data and   {\it it is the greatest}.
    \item $\mu_f$  is a capacity that represents the training data and {\it it is the lowest}.
    \item For any capacity  $\mu : 2^\mathcal{C} \rightarrow L$ that represents the training data, we have: 
$$\mu_f \leq \mu \leq \mu_e.$$
\end{itemize}
\end{corollary}

\noindent The methods presented in this article for obtaining the extremal capacities $\mu_e$ and $\mu_f$ that represent the training data, based on Sanchez's results \cite{sanchez1976resolution}, are simpler than those given in \cite{rico2005preference}. 

\subsection{Constructing \texorpdfstring{$q$-maxitives and $q$-minitives}{q-maxitives and q-minitives} capacities}

For a given training data item $(x^{(k)}, \alpha^{(k)})$, a necessary condition for being able to relate the object $x^{(k)}$ to its targeted global evaluation $\alpha^{(k)}$ using a Sugeno integral is: $$\min_{i \in \mathcal{C}} x_i^{(k)} \leq \alpha^{(k)} \leq \max_{i \in \mathcal{C}} x_i^{(k)}.$$ For our learning problem, if for all $k \in \{1,2,\dots,N\}$, each training data item $(x^{(k)}, \alpha^{(k)})$ satisfies this constraint, then we can provide sufficient conditions for the existence of $q$-maxitive and $q$-minitive capacities (where $1 \leq q \leq n$), which we deduce from the extremal solutions of the systems $(S)$ and $(\Sigma)$.

\begin{proposition}\label{propg:qmax}
Let $q\in \{1 , 2 , \dots n\}$. Assume that there exists     $A\subseteq \mathcal{C}$ such that $\mid A \mid = q$ and $e(A) = 1$. We put:
\begin{align*}
\mu_\vee &: 2^\mathcal{C} \rightarrow L \nonumber\\ 
 &: X \mapsto \mu_\vee(X) \nonumber\\
 &= \begin{cases}
0  & \text{if} \,\,\,\,\,\,   X = \emptyset\\
e(X)  & \text{if}\,\, 0 < \mid X \mid \leq q\\
\max\limits_{\emptyset \subset Y \subset X , \mid Y  \mid \leq q}\,e(Y) & \text{if}\mid X \mid > q
 \end{cases}.
\end{align*}

Then, $\mu_\vee$ is a $q$-maxitive capacity on $\mathcal{C}$.
\end{proposition}
\begin{proposition}\label{propg:qprmin}
Let $q\in \{1 , 2 , \dots n\}$. Assume that there exists  $  A'\subseteq \mathcal{C}$  such that $\mid A' \mid = n-q$  and  $f(A') = 0$. We put:
\begin{align*}
\mu_{\wedge} &: 2^\mathcal{C} \rightarrow L \nonumber\\ 
 &: X \mapsto \mu_{\wedge}(X)  \nonumber\\
 &= \begin{cases}
     1  & \text{if} \,\,\,\,\,\,\,\,\,\,\,\,X = \mathcal{C}\\
     f(X) & \text{if}  \,\,  n-q \leq \mid X \mid < n\\
  \min\limits_{Y \supset X , n-q \leq \mid Y \mid < n}\,f(Y) & \text{if} \,\,\,\,\,\,\,\, \mid X \mid < n-q
 \end{cases}.
\end{align*}
Then, $\mu_{\wedge}$ is a $q$-minitive capacity on $\mathcal{C}$.  
\end{proposition}
\subsection{Example}
Let $n = 3$ criteria i.e., $\mathcal{C} = \{ 1, 2, 3\}$. The scale is $L = \{ 0.05 \cdot l \mid 1 \leq l \leq 20 \}$. We consider the following $N  = 3$ training data items: 
    \begin{itemize}
        \item[-] $x^{(1)} = (0.15 , 0.2 , 0.3)$, $\alpha^{(1)} = 0.2$,
        \item[-] $x^{(2)} = (0.5 , 0.25 , 0.3)$, $\alpha^{(2)} = 0.3$,
        \item[-] $x^{(3)} = (0.4 , 0.7 , 0.35)$, $\alpha^{(3)} = 0.4$.
    \end{itemize}
\noindent We construct the system $(S): M \Box_{\min}^{\max} X = \alpha$ where:  
$$M = \begin{blockarray}{c|c|c|c|c|c|c}
 \{ 1 \} & \{ 2 \} & \{ 3 \} & \{1, 2\} & \{ 1,3\} & \{2,3\} & \{1,2,3\} \\ \hline 
 \begin{block}{c|c|c|c|c|c|c}
$0.15$&$0.2$&$0.3$&$0.15$&$0.15$&$0.2$&$0.15$ 
\\
$0.5$&$0.25$&$0.3$&$0.25$&$0.3$&$0.25$&$0.25$\\
$0.4$& $0.7$& $0.35$& $0.4$&$0.35$&$0.35$& $0.35$ \\
\end{block}
\end{blockarray} \text{ and } \alpha = \begin{bmatrix}
    0.2 \\
    0.3 \\
    0.4 
\end{bmatrix}.$$  The system $(S)$ is consistent and its greatest solution (\ref{eq:gr:sol}) is:
$$e = \begin{blockarray}{c|c}\begin{block}{c|c}
     \{ 1 \} & 0.3 \\ \hline  
     \{ 2 \} & 0.4 \\ \hline  
     \{ 3 \} & 0.2 \\\hline  
     \{1, 2\} & 1 \\\hline  
     \{ 1,3\} & 1 \\ \hline  
     \{2,3\} & 1 \\ \hline  
     \{1,2,3\} & 1\\\hline  
\end{block}
\end{blockarray}$$
and therefore 
we  have  $e(C) = 1$. The set function denoted by  
$\mu_e$, see (\ref{eq:mue}), is a 2-maxitive capacity which represents the data. It is the greatest, and it is not a possibility measure, as we have $ \mu_e(\{1, 2\}) = 1 >   \max(\mu_e(\{1\}) , \mu_e(\{2\})$.
We can similarly construct the system $(\Sigma)$ and obtain the capacity $\mu_f$ which represents the data and is the lowest capacity representing the training data.  

\section{Learning  \texorpdfstring{$q$}{q}-maxitive/minitive capacities in practice}
\label{sec:qmaxqmin}

\noindent In practice, we restrict ourselves to learning the measures of the sets of criteria with cardinality less than or equal to $q$ (with $1 \leq q < n$) of a capacity by considering the following system:
\[ (S_q): M_q \Box_{\min}^{\max} X = \alpha,\]
\noindent where $M_q = \begin{bmatrix}
    m_{k,A}
\end{bmatrix}_{1 \leq k \leq N, A \in 2^\mathcal{C}\backslash \{ \emptyset \}, \mid A \mid \leq q}$ and we denote by $e_q$  the potential greatest solution of  $(S_q)$,  see~(\ref{eq:gr:sol}). For $q = n$,  it is clear that the system  $(S_n)$ coincide  with the system $(S)$, see  (\ref{eq:sysLearnS}). 

\begin{property}\label{propri:maxitive}
 The system $(S_q)$ is said to verify the $q$-maxitivity property  if $\exists A \subseteq C$ such that  $0 < \mid A \mid = q$ and  $e_q(A) =1$. 
 \end{property}

  \noindent  
For any non-empty subset $ A \subseteq \mathcal C$ such that  $\mid A\mid \leq q$, we have $e_q(A) = e(A)$. The condition   $e_q(A) = 1$ means that 
  $\forall k \in\{1 , 2 , \dots , N\}\,,\, \min\limits_{i \in A}\, x_i^{(k)} \leq \alpha^{(k)}$, see  Lemma \ref{lemma:calculs}. 
We have the following result:

 \begin{theorem}\label{th:qmaxred}
The following two statements are equivalent:
 \begin{enumerate}
     \item There exists a   $q$-maxitive capacity   $\mu : 2^\mathcal{C} \rightarrow L$ which represents the training data:
$\forall k\in \{1 , 2 , \dots , N\}, \text{ we have } S_\mu(x^{(k)}) = \alpha^{(k)}$, 
\item The system $(S_q) : M_q \Box_{\min}^{\max} X = \alpha$ is consistent and it verifies  Property \ref{propri:maxitive}.
 \end{enumerate}
 \end{theorem}

\noindent
{\it Sketch of the proof:} Let $\mu : 2^{\mathcal C} \rightarrow L$ a $q$-maxitive capacity.  For all $A\in 2^{\mathcal C}\backslash\{\emptyset\}$ verifying $\mid A \mid \leq q$, we put $v(A) = \mu(A)$.
Therefore, it results from the formula (\ref{eq:sugenoformulas-maxmin}) that we have:

\[  \forall k\in\{1 , 2 , \dots , N\},\,\, S_\mu(x^{(k)}) = \alpha^{(k)} \Longleftrightarrow M_q\Box_{\min}^{\max} v = \alpha.
\]
\vskip0.3em 
For $q\in \{ 1 , 2 ,\dots , n\}$, the learning of $q$-minitive capacity in practice is analogous. We construct the reduced system:   
\begin{equation}\label{eq:sys:sigmaQ}
    (\Sigma_{q}) : \Gamma_{q} \Box_{\min}^{\max}  X = \alpha
\end{equation}
with  $  \Gamma_{q} = [\gamma_{k,A}]_{1 \leq k \leq N, A \in 2^\mathcal{C}\backslash\{\mathcal{C}\}, \mid A \mid\geq n-q}$ and we note $f_{q}$ its potential lowest solution, see  (\ref{eq:lt:sol}).

\begin{property}\label{propri:minitive}
     The system  $(\Sigma_{q})$ is said to verify the $q$-minitivity property 
 if $\exists A \subset  \mathcal C$ such that   $\mid A \mid = n- q$ and $f_{q}(A) =0$.\end{property}  

For any subset  $ A \subset  \mathcal C$ such that $\mid A \mid \geq n -q$, we have $f_q(A) = f(A)$. The condition   $f_q(A) = 0$ means that $\forall k \in\{1 , 2 , \dots , N\},\, \alpha^{(k)} \leq \max_{i \in \overline{A}}\, x_i^{(k)}$ (Lemma  \ref{lemma:calculs}). 

  \begin{theorem}\label{th:qmin}
The following two statements are equivalent:
 \begin{enumerate}
     \item There exists a $q$-minitive capacity $\mu : 2^\mathcal{C} \rightarrow L$ which represents the training data 
$\forall k\in \{1 , 2 , \dots , N\}, \text{we have } S_\mu(x^{(k)}) = \alpha^{(k)}$, 
\item The system $(\Sigma_{q}) : \Gamma_{q} \Box_{\min}^{\max} X = \alpha$ is consistent and it verifies Property \ref{propri:minitive}.
 \end{enumerate}
 \end{theorem}
{\it Sketch of the proof:} let $\mu : 2^{\mathcal C} \rightarrow L$ a $q$-minitive capacity.   For all $A\in 2^{\mathcal C}\backslash\{\mathcal C\}$ verifying 
$\mid A \mid \geq n -q$, we put $w(A) = \mu(A)$.
Therefore, it results from the formula (\ref{eq:sugenoformulas-minmax}) that we have:

\[ \forall k\in \{1 , 2 , \dots , N\},\,\, S_\mu(x^{(k)}) = \alpha^{(k)}\!  \Longleftrightarrow \!\Gamma_q\Box_{\max}^{\min} w = \alpha.\]

\subsection{Learning approximate capacities}

\noindent If we use the systems of fuzzy relational equations $(S_q)$ and $(\Sigma_{q})$ with real data, which may be subject to noise or contain outliers, the systems $(S_q)$ and $(\Sigma_{q})$ may become inconsistent. In this case, when  $L = [0, 1]$, we can rely on the work of \cite{baaj2024handling} on the handling of inconsistent systems of fuzzy relational equations to obtain capacities that approximately represent the training data. For an inconsistent system of max-min fuzzy relational equations $A \Box_{\min}^{\max} x = b$, the author of \cite{baaj2024handling} provides an explicit analytical formula for computing the Chebyshev distance, denoted by $\Delta = \Delta(A, b) = \min_{c \in \mathcal{C}} \Vert b - c \Vert_{\infty}$ (expressed with the $L_\infty$ norm), between the second member $b$ and the set
$$\mathcal{C} = \{ c \in [0,1]^{n \times 1} \mid \text{the system } A \Box_{\min}^{\max} x = c \text{ is consistent} \}$$
formed by the second members of the consistent systems defined with the same matrix: the matrix $A$ of the considered inconsistent system. The same result is provided in \cite{baaj2024handling} for systems of $\min-\max$ fuzzy relational equations, where the Chebyshev distance is denoted by $\nabla$. 
The formula of the Chebyshev distance  $\Delta_q = \Delta(M_q, \alpha)$ of the system $(S_q)$ is as follows (we use the notation $x^+ = \max(x,0)$):
\begin{equation}\label{eq:delta}
    \Delta_q = \max_{1 \leq i \leq N} \delta_i \quad \text{ with  } \quad \delta_i =     \min_{A \in 2^\mathcal{C}\backslash\{\emptyset\}, \mid A \mid \leq q} \delta_{i,A}
\end{equation}
\noindent where 
$\delta_{i,A}=  \max(\alpha^{(i)} - m_{i,A})^+, \max\limits_{1\leq l \leq N} \sigma_G(\alpha^{(i)} , m_{l,A},\alpha^{(l)})) \text{ and } 
\sigma_G(\alpha^{(i)} , m_{l,A},\alpha^{(l)}) = \min(\frac{(\alpha^{(i)} -  \alpha^{(l)})^+}{2}, (m_{l,A}-\alpha^{(l)})^+).$
The value of $\Delta_q$ gives an estimate of the quality of the training data used to build the system $(S_q)$, i.e., the data are of good quality if $\Delta_q \approx 0$ and not otherwise. Note that for $q \leq q'$, we have $\Delta_{q'} \leq \Delta_{q}$.
\vskip0.3em

The tools given in \cite{baaj2024handling} can then be used to obtain approximate solutions of the system $(S_q)$ via $\Delta_q$, from which $q$-maxitive capacities can be deduced.  
We use the greatest approximate solution of the system $(S_q)$, which is the vector $\eta_q  = M_q^t \Box_{\rightarrow_G}^{\min}\overline{\alpha}(\Delta_q)$ where $\overline{\alpha}(\Delta_q) = \begin{bmatrix}
    \min(\alpha^{(i)} + \Delta_q, 1)
\end{bmatrix}_{1 \leq i \leq N}$, see \cite{baaj2024handling}, to give sufficient conditions for the existence of the greatest approximate $q$-maxitive capacity, and we show how to construct it:
\begin{proposition}\label{mu*}
If $\exists A\subseteq \mathcal{C}$  such that $\mid A \mid = q$  and $\eta_q(A) = 1$, we put:
\begin{align*}
\mu^\ast &: 2^\mathcal{C} \rightarrow L \nonumber\\ 
 &: X \mapsto \mu^\ast(X) \nonumber\\
 &= \begin{cases}
0  & \text{if} \,\,\,\,\,\,   X = \emptyset\\
\eta_q(X)  & \text{if}\,\, 0 < \mid X \mid \leq q\\
\max\limits_{\emptyset \subset Y \subset X , \mid Y  \mid \leq q} \eta_q(Y) & \text{if}\mid X \mid > q
 \end{cases}.    
\end{align*}

Then $\mu^*$ is a $q$-maxitive capacity that verifies   $\max_{1 \leq k \leq N}\,
\mid S_{\mu^*}(x^{(k)}) - \alpha^{(k)} \mid  = \Delta_q$. 
\end{proposition}
The proof of this proposition relies on the fact that the vector $[S_{\mu^*}(x^{(k)})]_{1 \leq k \leq N}$ (resp. the vector $\eta_q$), 
is the greatest Chebyshev approximation (resp. the greatest approximate solution) of the system $(S_q) :   M_q \Box_{min}^{max} X = \alpha$, see \cite{baaj2024handling}. \\
\noindent
Using this proposition and the results of \cite{baaj2024handling}, we deduce the following two results:

\begin{theorem}\label{th:DeltaProp}
If $\exists A\subseteq \mathcal{C}$  such that $\mid A \mid = q$  and $\eta_q(A) = 1$ then:

\[ \min_{\text{ $q$-maxitive capacity } \mu} \max_{1 \leq k \leq N}\,
\mid S_\mu(x^{(k)}) - \alpha^{(k)} \mid = \Delta_q. \]
The minimal learning error of the $q$-maxitive capacities is equal to $\Delta_q$.
\end{theorem}
\noindent
The idea of the proof is based on Proposition \ref{mu*} and the fact that for any $q$-maxitive capacity $\mu$, the system $M_q \Box_{\min}^{\max} X = c$, where $c = [S_\mu(x^{(k)})]_{1 \leq k \leq N}$, is a consistent system. Note that the algorithm in \cite{beliakov2020robust} provides an approximate value of $ \min\limits_{\mu \text{ capacity }} \max_{1 \leq k \leq N}\, \mid S_\mu(x^{(k)}) - \alpha^{(k)} \mid$, whereas the formula (\ref{eq:delta}) allows us to obtain the exact value, i.e., $\Delta_n$. As a consequence of this theorem, we have the following important result:

\begin{corollary}\label{cor:greatestqmax}
 If $\exists A\subseteq \mathcal{C}$  such that $\mid A \mid = q$  and $\eta_q(A) = 1$,  then the capacity $\mu^*$ (Proposition \ref{mu*}), is the greatest 
  $q$-maxitive capacity   $\mu$ verifying $\max_{1 \leq k \leq N}\,
\mid S_\mu(x^{(k)}) - \alpha^{(k)} \mid  = \Delta_q$.
\end{corollary}
\noindent
To learn approximate $q$-minitive capacities, we use the system $(\Sigma_q) : \Gamma_q \Box_{\max}^{\min} X = \alpha$, see (\ref{eq:sys:sigmaQ}), whose Chebyshev distance is:$$\nabla_q = \max_{1 \leq i \leq N} \nabla_i \quad \text{ with } \quad \nabla_i = \min_{A \in 2^\mathcal{C}\backslash\{\mathcal{C}\}, \mid A \mid \geq n-q} \nabla_{i,A}$$
where $\nabla_{i,A}=  \max( (\gamma_{i,A}- \alpha^{(i)} )^+,   \max\limits_{1\leq l \leq N} \sigma_\epsilon(\alpha^{(i)},\gamma_{l,A},\alpha^{(l)}))$ and  
$\sigma_\epsilon(\alpha^{(i)},\gamma_{l,A},\alpha^{(l)}) =  \min(\frac{(  \alpha^{(l)} - \alpha^{(i)})^+}{2}, (\alpha^{(l)} - \gamma_{l,A})^+)$. We use the lowest approximate solution of the system $(\Sigma_q)$, which is the vector $\nu_q  = \Gamma_q^t \Box_{\epsilon}^{\max}\underline{\alpha}(\nabla_q)$ where $\underline{\alpha}(\nabla_q) = \begin{bmatrix}
    \max(\alpha^{(i)} - \nabla_q, 0)
\end{bmatrix}_{1 \leq i \leq N}$, see \cite{baaj2024handling}. As in the case of $q$-maxitive capacities, we have:
\begin{proposition}\label{mu_*}
If $\exists A\subseteq \mathcal{C}$  such that $\mid A \mid =  n - q$  and $\nu_q(A) = 0$,  we put: 
\begin{align*}
\mu_\ast &: 2^\mathcal{C} \rightarrow L \nonumber\\ 
 &: X \mapsto \mu_\ast(X) \nonumber\\
 &= \begin{cases}
1  & \text{si} \,\,\,\,\,\,   X = \mathcal C\\
\nu_q(X)  & \text{si}\,\, n - q\leq  \mid X \mid < n  \\
\min\limits_{    Y \supset X , n - q\leq\mid Y  \mid <n  } \nu_q(Y) & \text{si}\mid X \mid < n -q
 \end{cases}.
\end{align*}
Then,  $\mu_*$ is a $q$-minitive capacity which verifies   $\max_{1 \leq k \leq N}\,
\mid S_{\mu_*}(x^{(k)}) - \alpha^{(k)} \mid  = \nabla_q$. 
\end{proposition}
The proof of this proposition relies on the fact that the vector $[S_{\mu_*}(x^{(k)})]_{1 \leq k \leq N}$ (respectively the vector $\nu_q$), 
is the lowest Chebyshev approximation (respectively the lowest approximate solution) of the system $(\Sigma_q) :   \Gamma_q \Box_{\min}^{\max} X = \alpha$, see \cite{baaj2024handling}. 
Using this proposition and \cite{baaj2024handling}, we deduce the following two results:
\begin{theorem}\label{th:DeltaProp1}
If $\exists A\subseteq \mathcal{C}$ such that $\mid A \mid =  n - q$  and $\nu_q(A) = 0$  then:

\[ \min_{\text{$q$-minitive capacity } \mu}\,
\max_{1 \leq k \leq N}\,
\mid S_\mu(x^{(k)}) - \alpha^{(k)} \mid = \nabla_q. \]
The minimal learning error of the $q$-minitive capacities is equal to $\nabla_q$.
\end{theorem}

The idea of the proof is based on Proposition \ref{mu_*} and the fact that for any $q$-minitive capacity $\mu$, the system   
$\Gamma_q \Box_{\min}^{\max} X = c$, where $c = [S_\mu(x^{(k)})]_{1 \leq k \leq N}$, is a consistent system. We have the following result:
\begin{corollary}\label{cor:lowestqmin}
 If $\exists A\subseteq \mathcal{C}$  such that $\mid A \mid =  n - q$  and $\nu_q(A) = 0$, then the capacity $\mu_*$ (Proposition \ref{mu_*}), is the lowest $q$-minitive capacity   $\mu$ verifying $\max_{1 \leq k \leq N}\,
\mid S_\mu(x^{(k)}) - \alpha^{(k)} \mid  = \nabla_q$.    
\end{corollary}

\section{Conclusion}

In this article, we have introduced a method for learning capacities of Sugeno integrals according to training data based on systems of fuzzy relational equations. In particular, we have shown that this approach can learn approximate  $q$-maxitive and $q$-minitive capacities in a precise sense.

In perspective, we are currently working on the relationship between the systems $(S)$ and $(\Sigma)$ when they are inconsistent, in order to relate their respective approximate solution sets using the tools of \cite{baaj2024handling}.  Finally, we want to test our method on real data. 

\subsection*{Acknowledgements}

The author thanks Sébastien Destercke for useful discussions.
\bibliographystyle{abbrv}
\bibliography{references}

\end{document}